\documentclass[sigconf,nonacm]{acmart}

\usepackage{graphicx}
\usepackage{amsmath,amssymb,amsfonts}
\usepackage{textcomp}
\usepackage{xcolor}
\usepackage{graphicx, subcaption}
\usepackage[T1]{fontenc}
\usepackage[utf8]{inputenc}
\usepackage[greek,english]{babel}
\usepackage{CJKutf8}
\usepackage{multirow}
\usepackage{verbatim}
\usepackage{hhline}
\usepackage{caption}
\usepackage{bbm}
\usepackage{array}
\usepackage{makecell}
\usepackage{mathtools, nccmath}
\usepackage{wrapfig}
\usepackage[normalem]{ulem}

\DeclarePairedDelimiter{\norm}{\lVert}{\rVert}

\AtBeginDocument{%
  \providecommand\BibTeX{{%
    \normalfont B\kern-0.5em{\scshape i\kern-0.25em b}\kern-0.8em\TeX}}}

\begin{document}
\begin{CJK}{UTF8}{mj}

\title{S\textsuperscript{3}NAS: Fast NPU-aware Neural Architecture Search Methodology}

\author{Jaeseong Lee}
\affiliation{%
  \institution{Seoul National University}
}
\email{thnkinbtfly@iris.snu.ac.kr}

\author{Duseok Kang}
\affiliation{%
  \institution{Seoul National University}
}
\email{kangds0829@snu.ac.kr}

\author{Soonhoi Ha}
\authornote{Corresponding Author.}
\affiliation{%
  \institution{Seoul National University}
}
\email{sha@snu.ac.kr}

\begin{abstract}
As the application area of convolutional neural networks (CNN) is growing in embedded devices, it becomes popular to use a hardware CNN accelerator, called neural processing unit (NPU), to achieve higher performance per watt than CPUs or GPUs. Recently, automated neural architecture search (NAS) emerges as the default technique to find a state-of-the-art CNN architecture with higher accuracy than manually-designed architectures for image classification. In this paper, we present a fast NPU-aware NAS methodology, called S\textsuperscript{3}NAS, to find a CNN architecture with higher accuracy than the existing ones under a given latency constraint. It consists of three steps: supernet design, Single-Path NAS for fast architecture exploration, and scaling. %
To widen the search space of the supernet structure that consists of stages, we allow stages to have a different number of blocks and blocks to have parallel layers of different kernel sizes. %
For a fast neural architecture search, we apply a modified Single-Path NAS technique to the proposed supernet structure. In this step, we assume a shorter latency constraint than the required to reduce the search space and the search time. The last step is to scale up the network maximally within the latency constraint. 
For accurate latency estimation, an analytical latency estimator is devised, based on a cycle-level NPU simulator that runs an entire CNN considering the memory access overhead accurately. %
With the proposed methodology, we are able to find a network in 3 hours using TPUv3, which shows 82.72\% top-1 accuracy on ImageNet with 11.66 ms latency. Code are released at \url{https://github.com/cap-lab/S3NAS}
\end{abstract}

\maketitle

\section{Introduction}

As there are growing needs of deep learning applications based on convolutional neural network(CNN) in embedded systems, improving the accuracy of CNNs under a given set of constraints on latency and energy consumption has brought keen interest to researchers as a challenging problem in various areas. A popular hardware solution is to develop a hardware accelerator, called neural processing unit (NPU), that achieves higher performance per watt than CPUs or GPUs. %

For a given hardware platform, several software techniques have been proposed to accelerate CNNs by approximate computing since deep learning applications can tolerate a certain range of computation inaccuracy. Some examples in this software approach are filter pruning~\cite{li2016pruning}, quantization~\cite{park2017weighted}, low-rank approximation~\cite{kim2015compression}. 
Accelerating CNNs is helpful to improve the accuracy by running a more compute-intensive CNN with higher accuracy within a given time budget. 

On the other hand, various algorithmic solutions have been proposed to improve the CNN architecture by introducing new operations, optimizing the hyper-parameters, or searching for better network architecture. New operations such as depth-wise convolution(DWConv)~\cite{chollet2017xception} and mobile inverted bottleneck (MBConv)~\cite{sandler2018mobilenetv2} have been developed to replace the regular full convolution. Recently, automated neural architecture search (NAS) emerges as the default technique to find a CNN architecture with higher accuracy than manually-designed architectures, particularly image classification. 

\begin{figure}[t]
  \centering
  \includegraphics[width=0.8\linewidth]{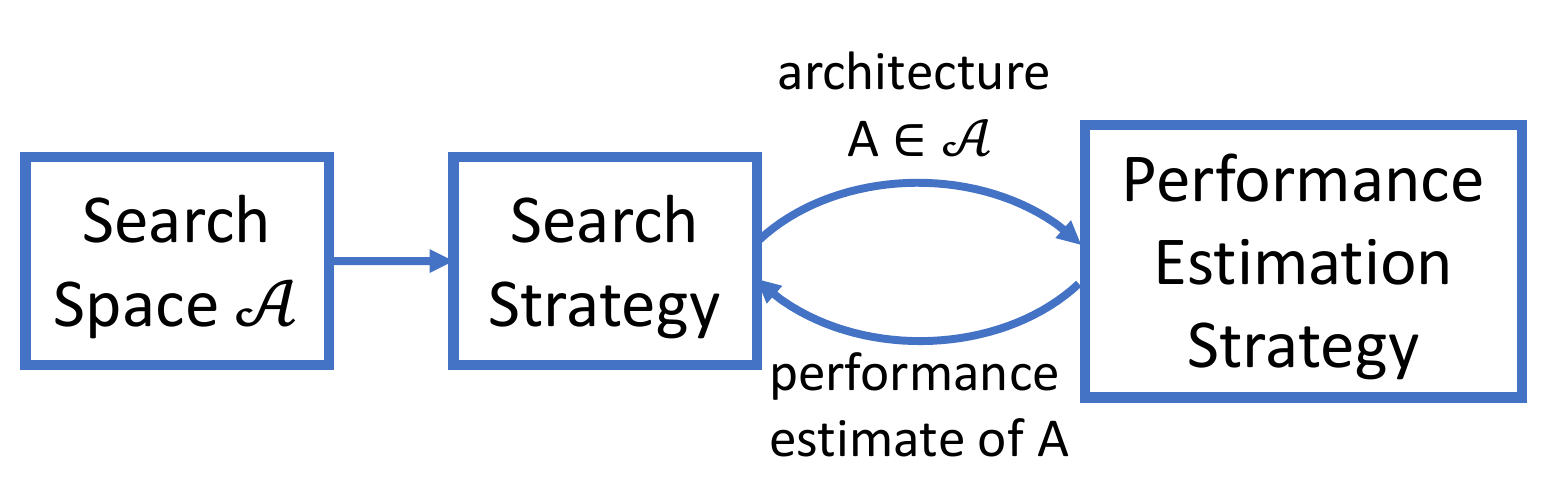}
  \caption{Typical procedure of NAS, adopted from~\cite{elsken2018neural}}
  \label{fig:NAS_explain}
\end{figure}

A NAS technique explores a predefined search space and estimates the performance for each candidate architecture to find an optimal one with the highest accuracy under a given latency constraint. Thus there are three factors that affect the performance of NAS, as shown in Figure~\ref{fig:NAS_explain}: search space, search strategy, and performance estimation. The search space of a NAS technique is usually restricted by a \textit{supernet} that defines the topology of the largest network to explore. Since the performance of a network depends on the hardware platform, the NAS technique needs to be customized to a given hardware platform. While numerous NAS techniques have been proposed with various search strategies recently, their assumed hardware platforms are mostly GPUs. In this paper, we present a customized NAS technique for an NPU, which produces a CNN architecture with a better accuracy-latency tradeoff than existing models. 

One of the most closely related work is the recently proposed NAS technique tailored for Google's Edge-TPU~\cite{gupta2020accelerator}. While %
MBConv is widely used for GPU-aware NAS techniques, they prefer to use a single full convolution by fusing expansion layer and DWConv layer
in some parts of the network, observing that the Edge-TPU runs the fused full convolution faster even though the required number of MAC (multiply-accumulate) operations is much larger. It confirms that the number of MAC operations is not a proper measure of latency, and platform-specific performance estimation is required. 

\begin{figure}
  \centering
  \includegraphics[width=1.0 \linewidth]{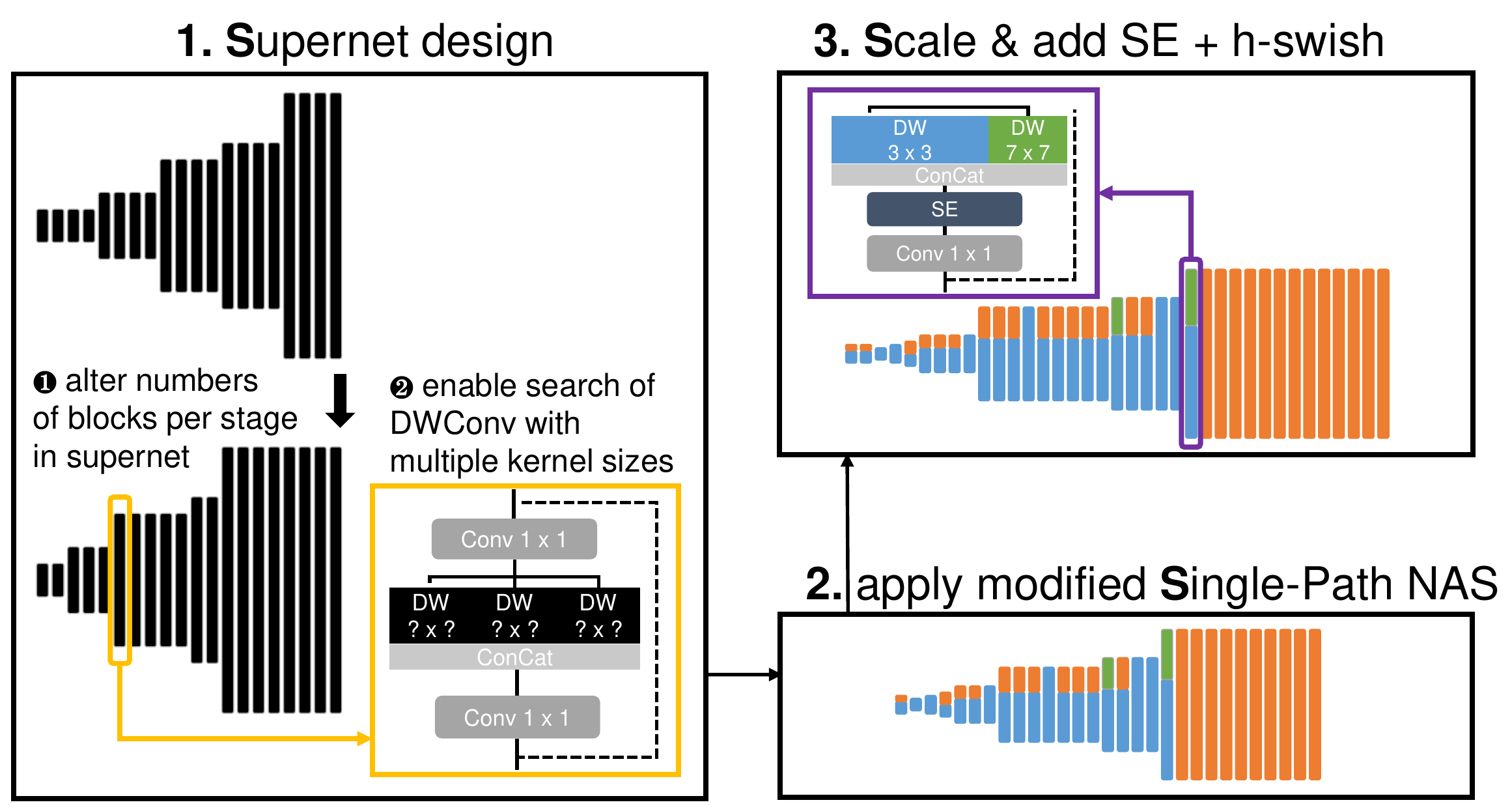}
  \caption{Overview of the proposed NAS technique.}
  \label{fig:overview}
\end{figure}

Since an NPU is much faster than a GPU, it enables us to explore the wider search space for NAS under a given latency constraint. Since there are many factors to define the search space, such as the number of layers, channels, kernel sizes, and so on, the search space grows exponentially as the allowed computation complexity grows. Hence, reducing the search space, as well as the search time, is very challenging for NPU-aware NAS techniques. While the aforementioned work for Google's Edge TPU trains each architecture candidate from scratch to estimate the performance, it is not computationally efficient. In contrast, we adopt a fast differentiable hardware-aware One-Shot NAS, called Single-Path NAS~\cite{stamoulis2019single}, in order to reduce the search time. 

Figure \ref{fig:overview} shows an overview of the proposed NAS methodology that consists of three steps. In the first step, we change the supernet structure of the Single-Path NAS, which has a hierarchical structure based on MobileNetV2~\cite{sandler2018mobilenetv2}: A supernet structure consists of a series of stages that contain a series of blocks containing an MBConv micro-architecture inside. 
Since the network accuracy depends on the supernet structure, we make two extensions on the supernet structure to widen the search space. First, we allow stages to have a different number of blocks, called depth of the stage, considering the effect of stage depth on the accuracy and the latency. Second, we add parallel layers with different kernel sizes in each block, adopting the idea of mixed depthwise convolution~\cite{tan2019mixconv} (MixConv). 

With the extended supernet structure, we apply the Single-Path NAS, which is also extended to support the extended supernet structure. In this step, we assume a shorter latency constraint than the required to reduce the search space and the search time. 
The last step is to 
scale up the baseline CNN adopting the compound %
scaling technique proposed in 
~\cite{tan2019efficientnet} until the latency constraint is met. The proposed NAS methodology is named as S\textsuperscript{3}NAS since it consists of 3 steps: \textbf{S}upernet design, \textbf{S}inglePath NAS, and \textbf{S}caling and post-processing.

For accurate latency estimation, an analytical latency estimator is devised, based on a cycle-level NPU simulator that runs an entire CNN considering the memory access overhead accurately. 
Since the NPU assumed in this paper can execute depth-wise separable convolution (DWConv), squeeze-and-excitation (SE), and h-swish activation function efficiently, the proposed supernet prefers DWConv to regular convolution. Observing that the accuracy is improved by around 1\% if SE and h-swish activation function are used, we add a post-processing phase after a CNN network is found by NAS to add SE layers and to replace ReLU to h-swish activation function.

Experiments show that the proposed NAS technique could improve the accuracy-latency tradeoff over existing SoTA CNN models. Our best model achieves \textbf{82.72\% top-1 accuracy on ImageNet with 11.66ms latency} without any special data augmentation. Note that the latency is estimated by cycle-accurate simulation. For a fair comparison with the related work, the latency of each compared network is also estimated with the same simulator. %

\section{Related Work}

\subsection{Neural Architecture Search (NAS)}
After an automated NAS technique based on reinforcement learning successfully found a better CNN architecture than manually-designed architectures~\cite{zoph2016neural}, extensive research has been conducted to develop various NAS techniques based on reinforcement learning~\cite{zoph2018learning,tan2019mnasnet}.
However, these NAS techniques are computationally intensive because they train each candidate architectures from scratch to estimate the goodness of it. Thus, one-shot neural architecture search approach~\cite{pham2018efficient} was introduced to reduce the search cost. 
In this approach, an over-parameterized super-model network is defined, and architecture search is performed by parameter optimization to reduce the complexity of the network. Gradient-based differentiable search has gained increasing popularity, and various NAS techniques have been proposed with different super-models and hyper-parameters~\cite{pham2018efficient,guo2019single,chu2019scarletnas,liu2018darts,cai2018proxylessnas}.

Among diverse techniques to decrease the search cost, Single-Path NAS~\cite{stamoulis2019single} was recently proposed to find a good architecture faster than the existing differentiable NAS techniques. This technique is extended to broaden the search space by including the squeeze-and-excitation (SE) block in the search space~\cite{stamoulis2020single}. Our work is grounded on the original Single-Path NAS technique.

\begin{figure*}[t]
  \centering
  \includegraphics[width=\linewidth]{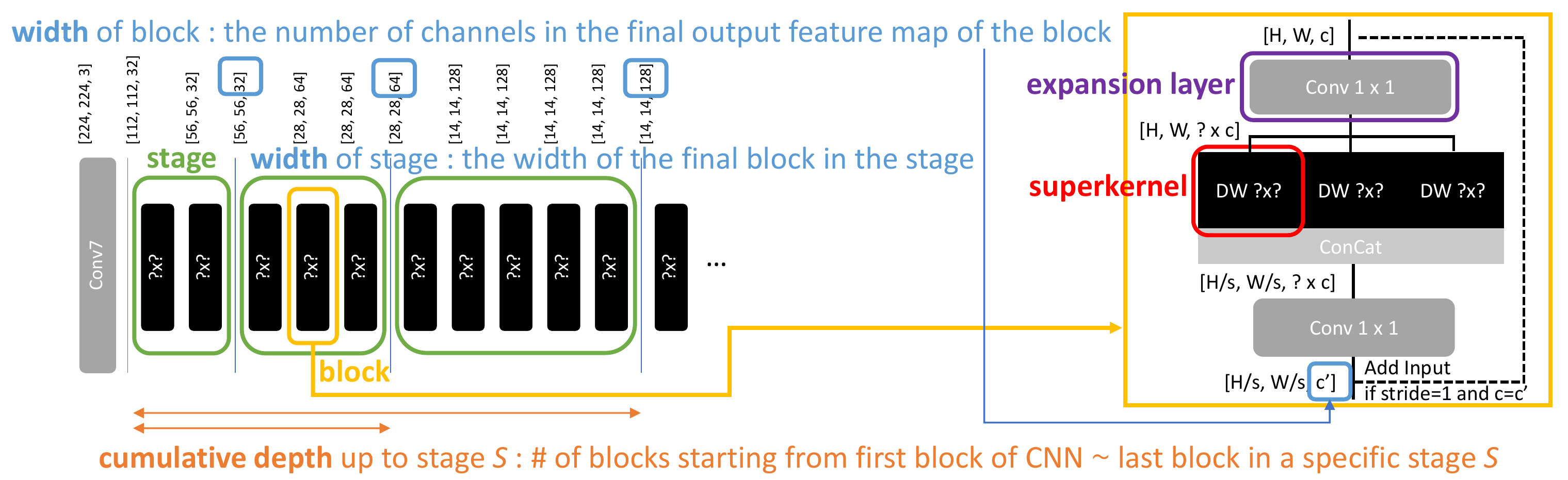}
  \caption{Definitions used in this paper. We depict our search space as an example.}
  \label{fig:definitions}
\end{figure*}

\subsection{Hardware-friendly Neural Architecture Design} 

Finding a hardware-friendly neural architecture has been facilitated as NAS algorithm improved. MNASNet~\cite{tan2019mnasnet} added a latency term in the objective function to discover better architectures with a given latency constraint on their target hardware platform. 
EfficientNet~\cite{tan2019efficientnet}, whose search method is similar to MNASNet, introduced a novel scaling method, called \textit{compound scaling}, to find more accurate networks as the latency constraint or FLOPS increases. Instead of finding a network directly for a given long latency constraint, they scale up the depth and the width of a small network with shorter latency and the input image size in a balanced way. They could achieve a set of networks with state-of-the-art performance over a range of latency constraints. %
They removed SE blocks and swish activation function from their search space for hardware platforms that do not support them efficiently to name the resultant network as EfficientNet-lite.

While EfficientNet searches a set of networks over a range of latency constraints by scaling up, Once-For-All~\cite{cai2019once} network takes an opposite approach, scaling down. They first train a super-graph architecture by a novel method called \textit{progressive shrinking} and search a sub-graph network that achieves good accuracy for a given latency constraint without re-training but cheap fine-tuning. They claim that a scaled-down network from the super-graph gives better accuracy than a network that is trained from scratch. They could find more accurate networks than EfficientNet for small latency constraints. %

To explore more efficient neural architectures on specific hardware, some NAS methods have proposed to define the design space of architecture exploration, tailored for the hardware platform. %
Gupta et al.~\cite{gupta2020accelerator} devised a building block named fused inverted bottleneck convolution block and showed that this block is often more efficient than MBConv on their target NPU, Edge-TPU. They adopted compound scaling method to find high-performing architectures on Edge-TPU.
Our work is closely related to this method. %
We devise a building block that consists of parallel DWConv layers with different kernel sizes, based on a preliminary experiment to find that it is better than the other alternative building blocks in terms of performance per latency~\cite{tan2019mixconv}. %
And we increase the search space by allowing stages to have a different number of blocks in the baseline supernet.

\subsection{Deciding the Depth of Stages}
A neural network typically consists of multiple stages, a sequence of blocks with the same number of output channels (width). There are studies on how to assign the number of blocks (depth) to each stage. Meng et al.~\cite{meng2020neural} observed that the way of assigning depth to each stage affects the accuracy. Moreover, they argued that the good depth assignment of each stage could be inherited from the shallow ones as the total depth is increased, and proposed a layer-growing NAS method that could significantly reduce the search space. Furthermore, Radosavovic et al.~\cite{radosavovic2020designing} discovered that among neural architectures with similar computational complexity, the ones whose stage width and depth have a quantized linear relationship tend to have higher accuracy. Based on similar observations, we apply this design principle to change the structure of the conventional One-Shot NAS supernet. In addition, we argue that placing more blocks in a stage with a larger width is beneficial. %

\subsection{Depthwise Convolution with Multiple Kernel Sizes}
While the original DWConv block uses a single kernel size for depthwise convolution, mixing multiple kernel sizes for depthwise convolution was recently proposed, named as \textit{MixConv}~\cite{tan2019mixconv}. Mixing multiple kernel sizes can be understood as having parallel branches inside a block. It is shown that MixConv is more efficient than ordinary DWConv~\cite{tan2019mixconv}. 
There exist some recent NAS methods~\cite{mei2019atomnas,chu2020mixpath} that also broaden their search space using DWConv with multiple kernel sizes to find better neural architectures. We adopt this approach in the supernet and  
formulate a differentiable latency model of this operation, enabling a latency-aware differentiable One-Shot NAS with MixConv.

\section{Background}

In this section, we will briefly review the Single-Path NAS technique and our target NPU.
Before going further, we define some terminologies used in this paper, as shown in Figure~\ref{fig:definitions}. A neural architecture consists of stages at the top level. A \textit{stage} consists of a sequence of blocks whose output feature maps have the same dimension. In the proposed supernet, a \textit{block} is defined as MBConv that typically starts with 1\texttimes1 conv (\textit{expansion layer}) and ends with 1\texttimes1 conv. %
Adopting the MixConv approach, the depthwise convolution layer consists of parallel \textit{superkernels} whose kernel size will be determined during the NAS process. The \textit{width of block} denotes the number of channels in the final output feature map of the block, and the \textit{width of stage} is the width of the final block in the stage. We will call the total number of blocks starting from the very first block in the network up to the last block in a specific stage S, as the \textit{cumulative depth up to stage} S.

\subsection{Single-Path NAS}\label{back_SPNAS}

Differentiable NAS methods usually define architecture parameters to choose which convolution layer to use in the block, training each convolution layer independently. %
Single-Path NAS~\cite{stamoulis2019single} reduce the search cost by decreasing the number of trainable parameters by sharing the kernel weights between convolution layers. 
The key idea is designing an over-parameterized depthwise convolution kernel named \textit{superkernel}, and letting each depthwise convolution kernel of candidate MBConvs directly inherit the weights of this superkernel.

\begin{figure}[!h]
  \centering
  \includegraphics[width=\linewidth]{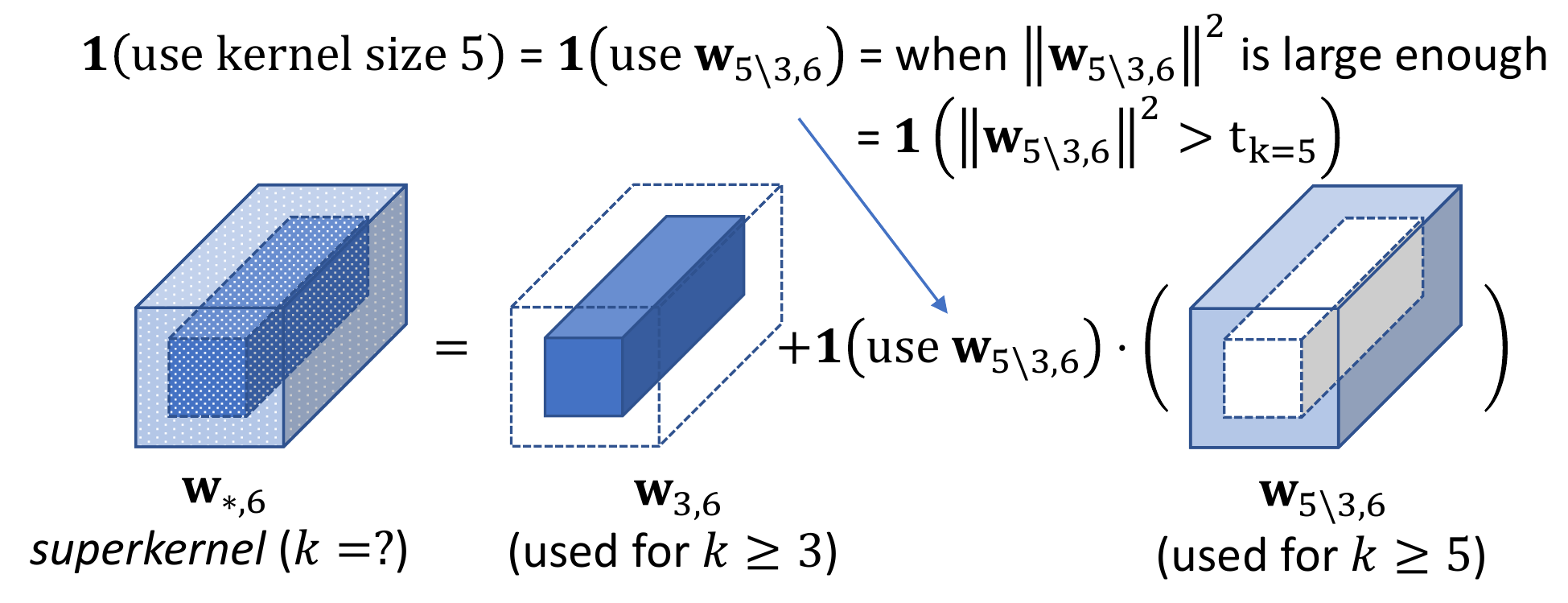}
  \caption{A searchable superkernel which can decide the kernel size.}
  \label{fig:SPNAS}
\end{figure}

Let $\mathbf{w}_{k,e}$ denote the depthwise convolution kernel of candidate MBConv with kernel size k and expansion ratio e (MBConv$_{k,e}$). First, they introduce a large $\mathbf{w}_{5,6}$, which is the DWConv kernel of MBConv$_{5,6}$. Then, the inner core of $\mathbf{w}_{5,6}$ can be considered as $\mathbf{w}_{3,6}$, a DWConv kernel of MBConv$_{3,6}$. 
A superkernel containing these two kernel size options can be expressed as Figure~\ref{fig:SPNAS}:
\begin{equation}\label{firstsuperkernel}
\mathbf{w}_{*,6} = \mathbf{w}_{3,6} + 
\mathbbm{1}(\rm{use~kernel~size~5}) \cdot
\mathbf{w}_{5\backslash3,6}
\end{equation}
where $\mathbf{w}_{5\backslash3, e}$ means the outer part, $\mathbf{w}_{5,e}-\mathbf{w}_{3,e}$.
Next, they formulate conditions to determine the kernel size. They define a certain threshold value $t$ and compare the norm of the kernel weights with the threshold. If the norm of a subset weight is larger than the threshold, it remains in the supernet. %
To this end, Eq.~(\ref{firstsuperkernel}) is changed as follows:
\begin{equation}
\mathbf{w}_{*,6} (t_{k=5}) = \mathbf{w}_{3,6} + 
\mathbbm{1}(\norm{\mathbf{w}_{5\backslash3, 6}}^2 > t_{k=5} ) \cdot
\mathbf{w}_{5\backslash3,6}
\end{equation}

The threshold value is also trainable to be automatically chosen during training. To enable back-propagation, they relax $\mathbbm{1} (x > t)$ to $\sigma(x-t)$ when computing gradients. In addition, they optimize kernel weights and threshold values simultaneously. For a given tight search time, this method is shown to be more effective than the other methods%
~\cite{stamoulis2020single}.

Moreover, we can vary the number of channels by varying the expansion ratio of each block: we can use only the first half channels of $\mathbf{w}_{5,6}$ and $\mathbf{w}_{3,6}$  as $\mathbf{w}_{5,3}$ and $\mathbf{w}_{3,3}$, respectively. By defining another set of trainable thresholds, the following formula is defined to determine the expansion ratio: %
\begin{multline} \label{combinedkern}
\mathbf{w}_{*, *} (t_{e=3}, t_{e=6}, t_{k=5}) = \mathbbm{1}(\norm{\mathbf{w}_{*, 3} (t_{k=5})}^2 > t_{e=3} ) \cdot \mathbf{w}_{*, 3} (t_{k=5}) + \\
\mathbbm{1}(\norm{\mathbf{w}_{*, 3} (t_{k=5})}^2 > t_{e=3} ) \cdot \mathbbm{1}(\norm{\mathbf{w}_{*, 6\backslash3} (t_{k=5})}^2 > t_{e=6} ) 
\cdot \mathbf{w}_{*, 6 \backslash 3} (t_{k=5})
\end{multline}
where $\mathbf{w}_{k, 6\backslash3}$ means the remaining half of channels, $\mathbf{w}_{k,6}-\mathbf{w}_{k,3}$.  Note that if $t_{e=3}$ is sufficiently large, all channels can be removed to make the block a plain skip connection. %
Thus, they replace the original depthwise convolution kernel of MBConv$_{5,6}$ with $\mathbf{w}_{*,*}$, yielding a differentiable and searchable MBConv with respect to the kernel size and expansion ratio.

They also design a differentiable latency-aware loss function to consider hardware latency in the search algorithm. 
To this end, they define a function to estimate latency as follows:

\begin{equation}
\begin{split}
    L^l_e = &
    \mathbbm{1}(\norm{\mathbf{w}_{*, 3} }^2 > t_{e=3} ) \cdot (P^l_{5,3} + \\ &\mathbbm{1}(\norm{\mathbf{w}_{*, 6\backslash3} }^2 > t_{e=6} ) 
    \cdot (P^l_{5,6} - P^l_{5,3}))
\end{split}
\end{equation}

\begin{equation}\label{orig_latformula}
\begin{split}
    L^l = &P^l_{3,6}/P^l_{5,6} \cdot L^l_e + \\
    &\mathbbm{1}(\norm{\mathbf{w}_{5\backslash3, 6}}^2 > t_{k=5} ) \cdot L^l_e \cdot (1 - P^l_{3,6}/P^l_{5,6})
\end{split}
\end{equation}
where $P^l_{k,e}$ is a profiled latency value for MBConv$_{k,e}$ for the $l$th block in the supernet. Note that they used $P^l_{3,6}$, $P^l_{5,3}$, and $P^l_{5,6}$ only to formulate $L^l$, and the latency for MBConv$_{3,3}$ is approximated using these values. Here is the latency-aware loss function designed:

\begin{equation}\label{origlossftn}
CE + \lambda \cdot log( \sum_{l} L^l)
\end{equation}
Finally, they search for a neural architecture in two phases. First, they train the supernet by randomly choosing one of the candidate subgraphs in each training step. In this phase, they use CrossEntropy loss only. Next, they enable latency-aware loss function and train the supernet with the loss function, to decide the threshold values. By doing this, they could get a high-quality neural architecture with only eight epochs of ImageNet training set.\footnote{In our implementation, we changed the probability of selecting each candidate MBConvs to be equal.}

\subsection{Target NPU}

Even though the proposed methodology can be applied to any type of NPU, the current implementation is made for an adder-tree type NPU, called MIDAP~\cite{kang2019novel}.
It has a fully-pipelined micro-architecture that consists of separate hardware modules and memory modules for convolution, activation function, and various reduction operations. Since it enables us to make a fully static schedule of operations without resource contention in the data path, we can estimate the end-to-end latency of a CNN quite accurately analytically. Unexpected delay may incur from off-chip DRAM delay that is not fully hidden by double buffering.

Another good feature of MIDAP is that it efficiently supports %
the following operations that would lower the MAC (multiply-accumulate) utilization in other NPUs that have many MAC units: pooling, DWConv, and  squeeze-and-excitation (SE). For DWConv operation, it does not use an adder tree but an alternative hardware logic that consists of a set of individual accumulators connected to the multiply units. For pooling and SE operations, reduction logic is included in the pipeline. 
Note that MIDAP has not been implemented as a real hardware chip yet but as a virtual prototype with a cycle-accurate simulator. Thanks to the cycle-accurate simulator that considers the DRAM access contention and parametrized DRAM access delay, we could build an accurate analytical model for end-to-end latency estimation, based on the profiling result with the simulator. %

Inverted bottleneck with depth-wise convolution (MBConv)~\cite{sandler2018mobilenetv2} is a popular building block in recent mobile-friendly networks. However, it is not efficiently supported in existing NPUs that do not have specialized hardware units for DWConv~\cite{gholami2018squeezenext, gupta2020accelerator}. Thus Gupta et al.~\cite{gupta2020accelerator} replaced an MBConv block with a fused building block that fuses an expansion layer and DWConv in MBConv into a single full convolution. Even though the fused block increases the number of multiplications significantly, it improves the MAC utilization larger so that the fused block is observed faster than MBConv on their target NPU, EdgeTPU. By adding this building block to their search space, they could successfully obtain different neural architectures for EdgeTPU from those for GPUs. 

Since DWConv is efficiently supported in MIDAP, however, the improvement of MAC utilization by fusing does not outweigh the increased computation complexity, which is observed in preliminary experiments. The experiment setup is similar to main experiment setup that will be explained in section~\ref{exp_proposed}. The experimental result is shown in Table~\ref{tab:FusedvsMB}. The latency constraint for fused block experiment is set to 7.0ms, while others are set to 2.15ms. In the combined experiment, we use the fused block in the 1st and the 2nd stages, and MBConv for the remaining stages since the latency gap between two building blocks is too high. As shown in the table, MBConv block shows the best tradeoff between accuracy and latency. Hence we prefer MBConv to the fused building block as the basic building block in the supernet for MIDAP.

\begin{table}[!ht]
\centering
  \caption{%
  NAS comparison with different building blocks}
  \label{tab:FusedvsMB}
  \begin{tabular}{ccc}
    \toprule
    building block&accuracy (\%)&latency (ms)\\
    \midrule
    Fused inverted bottleneck conv & 77.34 & 6.90\\
    \textbf{MBConv} & \textbf{77.75} & 2.05 \\
    Combined & 76.55 & 2.20\\
  \bottomrule
\end{tabular}
\end{table}

\section{Proposed S\textsuperscript{3}NAS Methodology}
\label{sec:propoesd}

In this section, we explain the proposed S\textsuperscript{3}NAS methodology that consists of three steps as displayed in Figure~\ref{fig:overview}. 

\subsection{Supernet Design}

The number of blocks is one of the key parameters in neural networks. It is observed that the total number of blocks affects the accuracy of neural architecture~\cite{he2016deep,tan2019efficientnet}. 
In conventional One-Shot NAS methods, each stage in the supernet has the same number of blocks~\cite{cai2018proxylessnas,stamoulis2019single,wu2019fbnet}.
On the other hand, some recent studies~\cite{meng2020neural,radosavovic2020designing} report that the way of assigning the number of blocks in each stage has a noticeable impact on the accuracy, even with the same number of blocks in total. Hence we allow stages in the supernet to have a different number of blocks.

\begin{figure}[!ht]
  \centering
  \includegraphics[width=\linewidth]{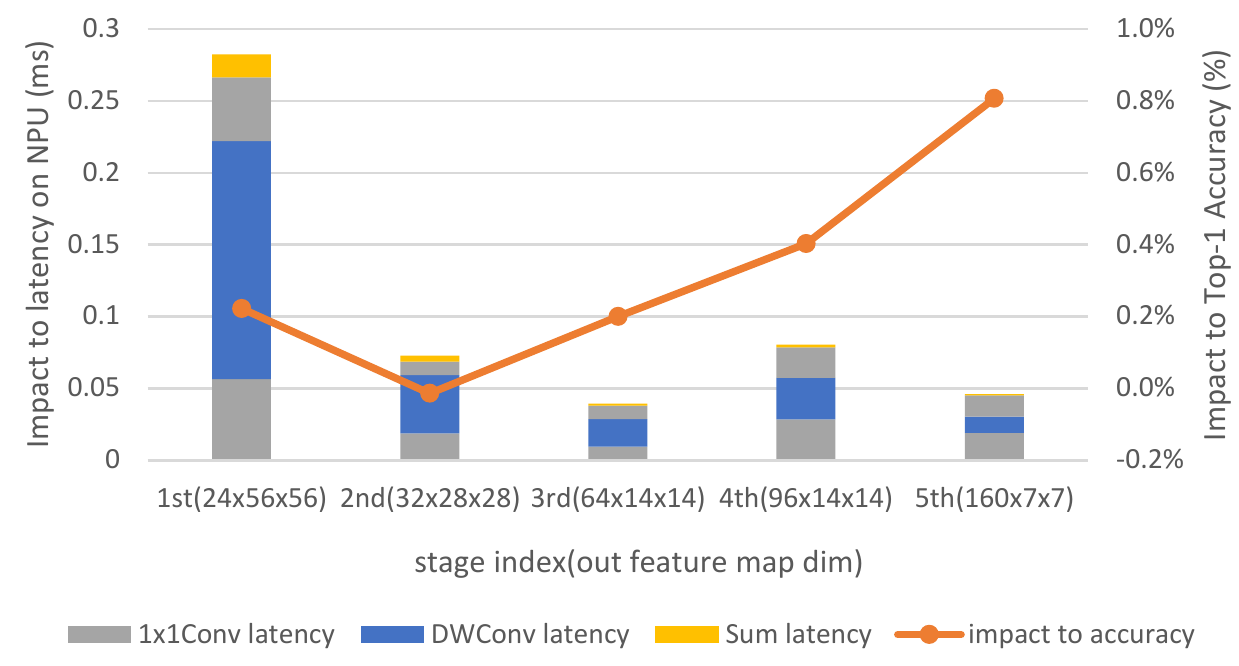}
  \caption{Impact of reducing the number of blocks in different stages in a MobileNetV2-based model}
  \label{fig:impactofblocks}
\end{figure}

We investigate the impact of assigning the number of blocks in the supernet with another preliminary experiment. %
We construct a network based on MobileNetV2, which has four blocks in every stage, and observe the change of accuracy as we reduce two blocks in a different stage in each experiment. %
Figure~\ref{fig:impactofblocks} shows that MBConvs with larger width has more impact on accuracy. 

As the number of multiplications in a DWConv is $W\times H\times C\times K^2$, the later stage of DWConv tends to have shorter latency since the reduction of $H\times W$ is larger than the increase of $C$. Thus the impact on the latency by increasing the number of blocks in a later stage is not significant as displayed in Figure~\ref{fig:impactofblocks}.

Thus, we place more blocks to stages with larger width in the supernet, %
making the cumulative depth up to a specific stage is proportional to the width of the stage, which is similar to PyramidNet~\cite{han2017deep}. A recent study~\cite{radosavovic2020designing} also claims that neural architectures with a linear relationship between the cumulative depth and the width tend to have higher accuracy with a similar amount of computation complexity. Our experiment shows that our modification to supernet enhances the efficiency of the search result in terms of accuracy as well as latency (Table~\ref{tab:depthwidth}).

Another feature of the proposed supernet is to use mixed convolution (MixConv) that mixes different kernel sizes in the depth-wise convolution layer~\cite{tan2019mixconv}. 
Some recent NAS methods~\cite{mei2019atomnas,chu2020mixpath} also broaden their search space using DWConv with various kernel sizes and could find better neural architectures. %

\begin{figure}[!ht]
  \includegraphics[width=0.45\linewidth]{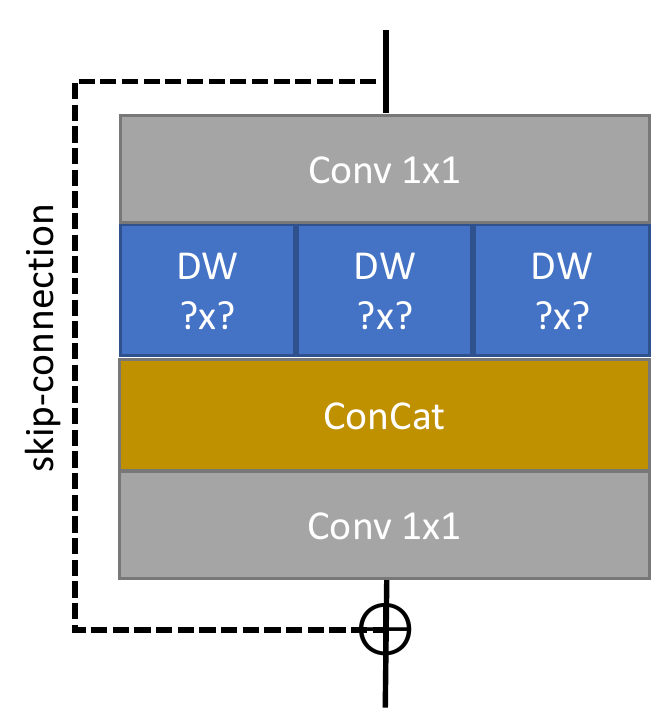}
  \caption{Our MixConv-based building block for supernet}
  \label{fig:superblock}
\end{figure}

Figure~\ref{fig:superblock} depicts our building block structure. This block starts and ends with 1\texttimes1 convolution, with $N$ searchable superkernels in the middle. Each searchable superkernel is designed similarly to Eq.~(\ref{combinedkern}), while we may use different threshold values in each superkernel. The kernel sizes and expansion ratios are selected among predetermined values. If the $j$-th searchable superkernel chooses an expansion ratio $e_j$, the $j$-th kernel has $e_j$  times more channels than the first 1\texttimes1 convolution. Compared with the original MixConv suggested in~\cite{tan2019mixconv}, the proposed building block supports more diverse combinations of kernel sizes and expansion ratios. %
It enhances the efficiency of search results on our target NPU (Table~\ref{tab:MBvsMBConv}).

We finish this subsection by highlighting the merit of Single-Path NAS on building a MixConv-based differentiable NAS. Conventional multi-path NAS methods would have difficulties when adding inverted bottleneck convolution with MixConv to their search space. Since the number of possible choices of such blocks grows proportionally to the partition number, multi-path NAS methods would introduce a significant increase in memory requirements and the search time. On the contrary, MixConv can be efficiently supported in Single-Path NAS, as explained below. %

\subsection{Modified SinglePath NAS}

We use a different latency estimation model, and a loss formula from the original SinglePath NAS technique explained in section~\ref{back_SPNAS}.

\subsubsection{Differentiable Latency Model}

Suppose we concatenate $N$ searchable superkernels to build a MixConv-based building block, and let $\vec{k}=(k_1, \cdots , k_N), \vec{e}=(e_1, \cdots , e_N)$ where $k_j, e_j$ denote the kernel size and the expansion ratio of the $j$th searchable superkernel. The estimated latency of a DWConv operation depends on the kernel size and the expansion ratio.

For latency formulation, we first define two condition variables, $F_{j,k_j}$ and $G_{j,e_j}$, that denote whether the $j$th searchable superkernel chooses the kernel size $k_j$ and the expansion ratio $e_j$, respectively; For example, $F_{j,k_j}$ is 1 if and only if the $j$th searchable superkernel chooses $k_j$, and 0 otherwise.

Let $\kappa_{1} <\cdots <\kappa_{K}$ be the candidate kernel sizes, and $0 = \epsilon_{1} < \cdots <\epsilon_{E}$ denote the candidate expansion ratios of the $j$th searchable superkernel, respectively.
Suppose $k_j=\kappa_{c}$, then $F_{j,k_j}$ can be formulated as follows: %

\begin{equation}\label{modified_kernel_cond}
\begin{split}
F_{j,k_j} &= \left( \prod_{2\leq i\leq c} \mathbbm{1}(\norm{\mathbf{w}_{j,\kappa_{i}\backslash\kappa_{i-1},\epsilon_{E}}}^2> t_{j,\kappa_i}) \right) \cdot f_{j, k_j} \text{,    where}\\
f_{j,k_j} &= 
\begin{cases}
    \mathbbm{1}(\norm{\mathbf{w}_{j,\kappa_{c+1}\backslash\kappa_{c},\epsilon_{E}}}^2<t_{j,\kappa_{c+1}}), & \text{if } c<K\\
    1,& \text{if } c=K
\end{cases}
\end{split}
\end{equation}

\begin{figure}[!ht]
  \includegraphics[width=\linewidth]{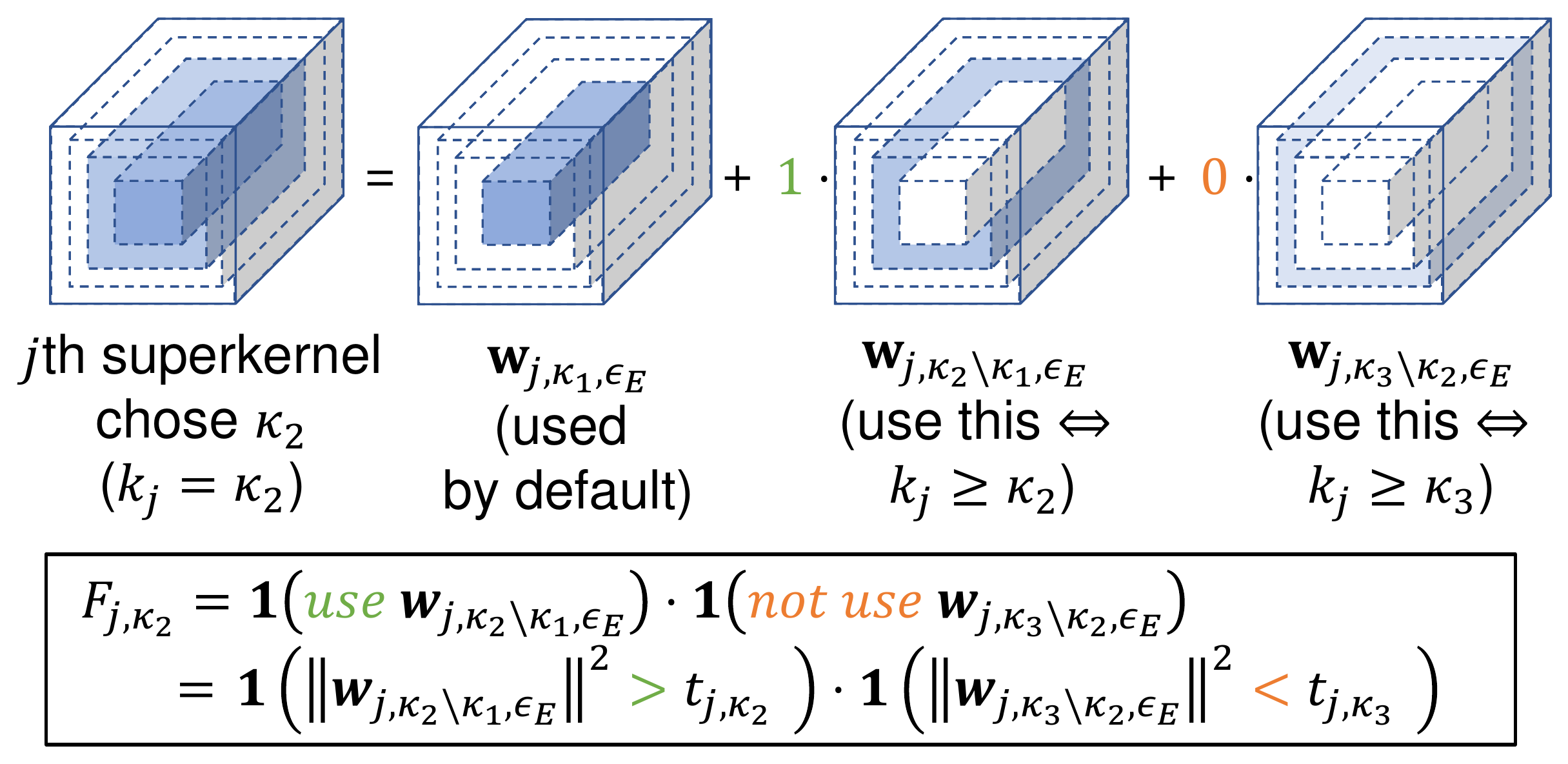}
  \caption{An example of Eq. (\ref{modified_kernel_cond}) when a searchable superkernel chose kernel size $\kappa_{2}$.}
  \label{fig:lat_formula_ex}
\end{figure}

Figure~\ref{fig:lat_formula_ex} depicts an example of this formula when the $j$th searchable superkernel that has four candidate kernel sizes $\kappa_{1}<\cdots<\kappa_{4}$ chooses $\kappa_{2}$ as the kernel size: $k_j=\kappa_{2}$. It means that weight $\mathbf{w}_{j,\kappa_{1},\epsilon_{E}}$ and $\mathbf{w}_{j,\kappa_{2}\backslash\kappa_{1},\epsilon_{E}}$ are used, but the remaining weights starting from $\mathbf{w}_{j,\kappa_{3}\backslash\kappa_{2},\epsilon_{E}}$ are not used.
Since $\mathbf{w}_{j,\kappa_{1},\epsilon_{E}}$ is always used, it is not included in the formula. To use $\mathbf{w}_{j,\kappa_{2}\backslash\kappa_{1},\epsilon_{E}}$, the norm of it has to be larger than $t_{j,\kappa_{2}}$ while the norm of $\mathbf{w}_{j,\kappa_{3}\backslash\kappa_{2},\epsilon_{E}}$ should not be larger than $t_{j,\kappa_{3}}$ to avoid the use of larger kernel sizes.

We can formulate $G_{j,e_j}$ similarly: 
\begin{align*}
G_{j,e_j} &= \left( \prod_{2\leq i\leq d} \mathbbm{1}(\norm{\mathbf{w}_{j,*,\epsilon_{i}\backslash\epsilon_{i-1}}}^2> t_{j,\epsilon_i}) \right) \cdot g_{j, e_j} \text{,    where}\\
g_{j,e_j} &= 
\begin{cases}
    \mathbbm{1}(\norm{\mathbf{w}_{j,*,\epsilon_{d+1}\backslash\epsilon_{d}}}^2<t_{j,\epsilon_{d+1}}), & \text{if } d<E\\
    1,& \text{if } d=E
\end{cases}
\end{align*}
when $e_j=\epsilon_{d}$. Then the condition for a MixConv-based building block to choose $\vec{k},\vec{e}$ can be expressed as $\prod_j^N F_{j,k_j} G_{j,e_j}$.

Now, the estimated latency of a single block is formulated as follows:
\begin{equation}\label{general_latestim}
L = \sum_{\vec{k},\vec{e}} (P(\vec{k},\vec{e}) \prod_j^N F_{j,k_j} G_{j,e_j})
\end{equation}
where $P(\vec{k}, \vec{e})$ denotes the profiled latency value of a MixConv-based building block corresponding to $\vec{k}, \vec{e}$.

Unlike the original Single-Path NAS that approximates the latency in Eq. (\ref{orig_latformula}) in some cases, we use the profiled latency value in all cases. %
Note that an expansion ratio can be zero%
, and if only one superkernel has a nonzero expansion ratio, the MixConv block is reduced to a plain MBConv block.
Finally, we can estimate the latency by summing up these estimated latencies for all superkernels in the block, $\sum L$.

\begin{figure}[!ht]
  \includegraphics[width=0.9\linewidth]{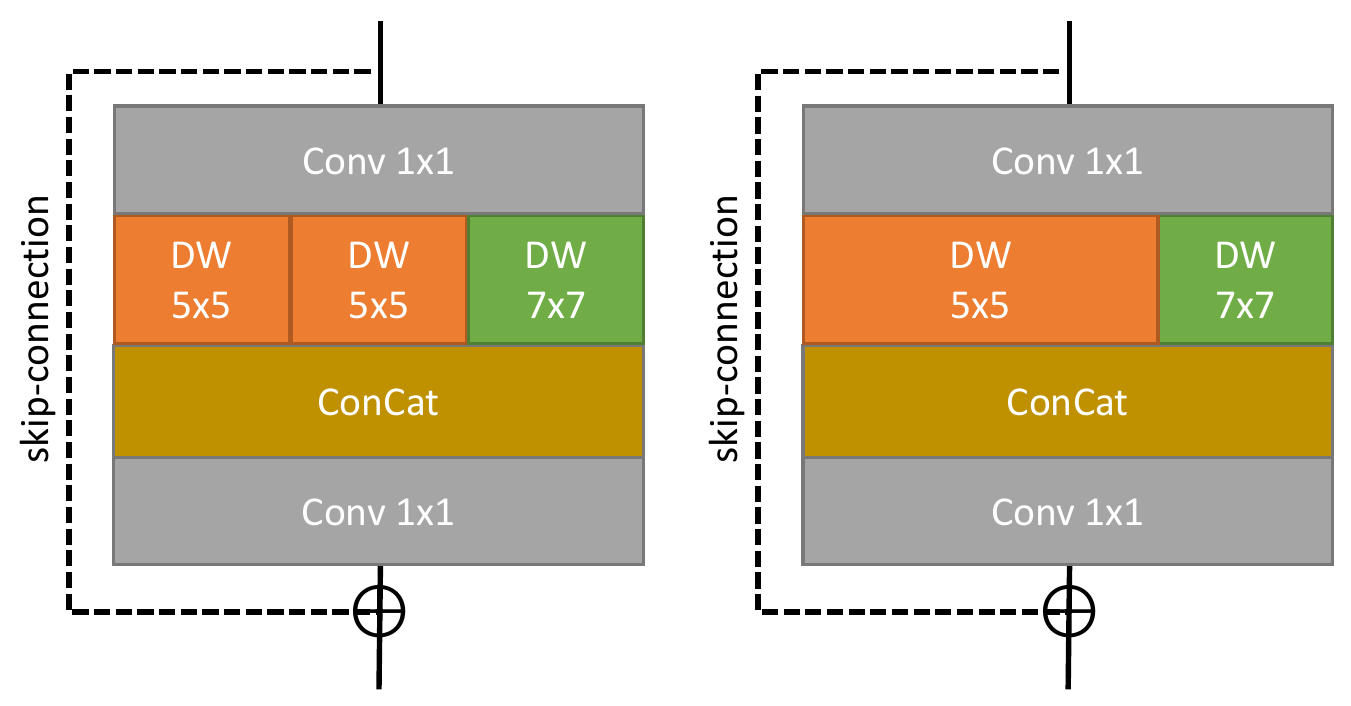}
  \caption{Two implementations of a same block. While their FLOPS are identical, their estimated latencies are different.}
  \label{fig:equivblock}
\end{figure}

Since each superkernel is treated independently, some superkernels may have the same kernel size and expansion ratio.  %
Then, even if two superkernel configurations express an equivalent block, as illustrated in Figure~\ref{fig:equivblock}, they may have different estimated latency values, which is an artifact of the proposed profiling-based latency estimation method. To avoid this artifact, we enforce that there is only one kernel for each kernel size in the MixConv block. That is, we merge two kernels of the same size into one; For instance, the left MixConv is translated to the right MixConv in Figure~\ref{fig:equivblock} before latency estimation.

\begin{figure}[!ht]
  \centering
  \includegraphics[width=0.8\linewidth]{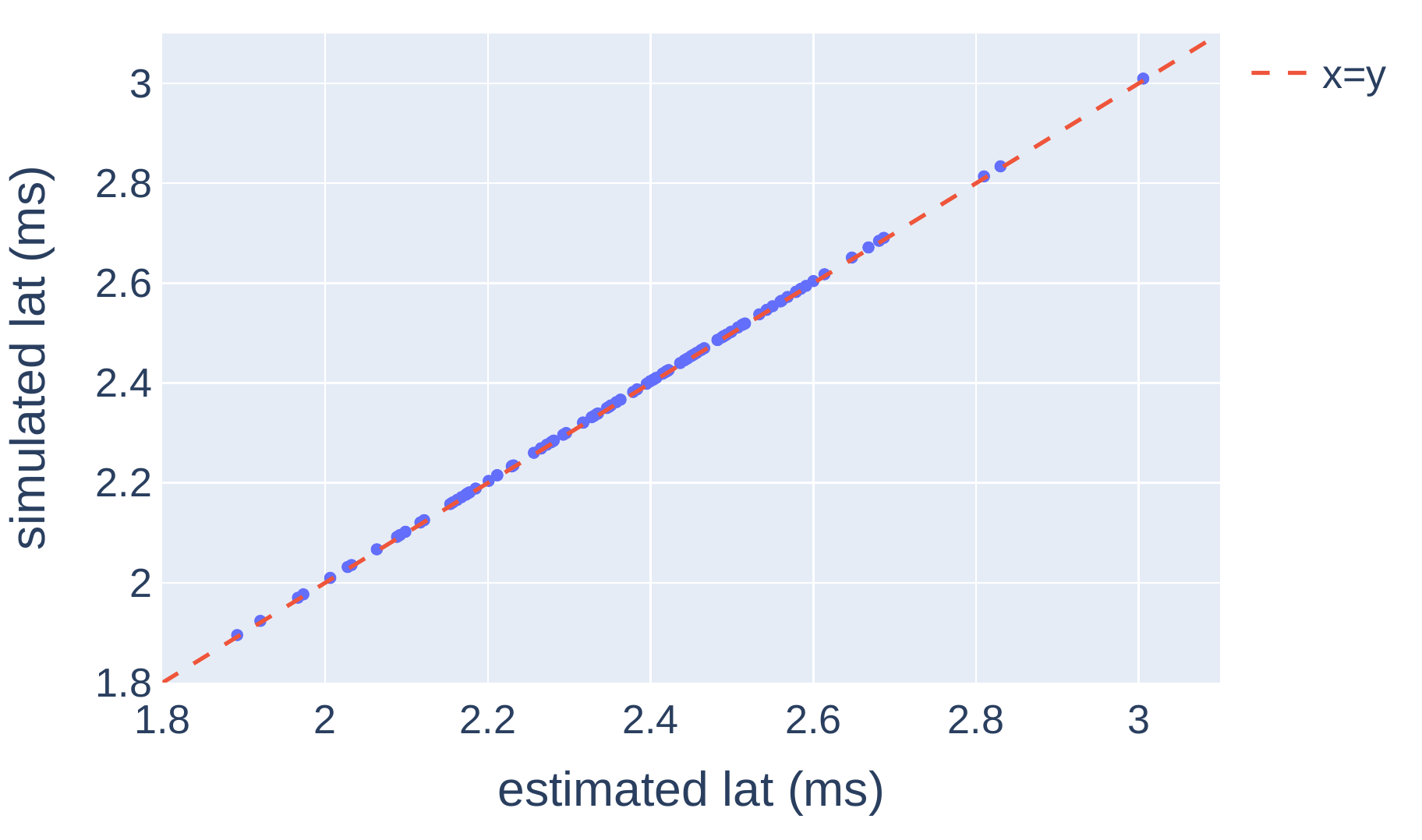}
  \caption{Accuracy of the proposed latency estimation model: MAPE is about 0.16\%.}
  \label{fig:latestim}
\end{figure}

Figure~\ref{fig:latestim} shows the estimated latency and simulated latency of randomly generated 100 models on our search space. It validates the accuracy of the proposed latency model, whose mean absolute percentage error(MAPE) is about 0.16\%.

\subsubsection{Differentiable Loss with Target Latency}

The existing hardware-aware differentiable NAS methods mostly define some hyperparameters to balance between accuracy and latency, including SinglePath NAS, whose loss function is defined as Eq.~(\ref{origlossftn}). Since there is no information on the target latency in the loss function, in case there is a strict latency constraint, they have to pay additional search costs for the hyperparameters to let the final architecture have no larger latency than the constraint. In addition, this process needs to be repeated whenever the target latency is changed. %

We propose to modify the loss function to %
activate the latency-aware loss term only when the estimated latency is larger than the latency constraint as follows:
\begin{equation} \label{lossformula}
CE + \lambda_1 \cdot log(1 + \lambda_2\cdot ReLU((\sum L) - T))
\end{equation}
Although this is not a panacea, this modification significantly eases the search process, which will be discussed in section~\ref{exp_proposed} with various experiments.

\subsection{Post-processing}

In the second step, we intentionally use shorter latency to reduce the search space for the baseline network. After finding the baseline network with a shorter latency, we apply compound scaling to find an architecture with the final latency constraint. In this step, we conduct post-processing to add SE block and h-swish activation function if beneficial. 

\subsubsection{Compound Scaling}
It is well known that increasing depth~\cite{he2016deep}, width~\cite{zagoruyko2016wide}, or input image size improves accuracy while it increases latency. However, if only one of these three factors is increased, the accuracy improvement is quickly saturated. Observing this fact, Tan et al.~\cite{tan2019efficientnet} proposed a \textit{compound scaling} method that increases all three factors together. A scaling coefficient is defined for each factor.
By judiciously assigning the scaling coefficients in a balanced fashion, they could improve the accuracy much larger than scaling a single factor only. 
Adopting this approach, we apply the compound scaling to the baseline architecture obtained in the previous step. Based on the ratio between the true latency constraint and the assumed latency constraint in the second step, we find the scaling coefficients considering the estimated latency increment. %
To keep the linear relationship between the width and cumulative depth, we use the same scaling coefficient for width and depth, differently from~\cite{tan2019efficientnet}. 
Note that how to realize scaling depends on the baseline architecture. While the baseline architecture assumed in~\cite{tan2019efficientnet} has a series of identical blocks in each stage, a stage consists of heterogeneous blocks in our baseline architecture. Thus depth scaling is not realized by merely adding new blocks in each stage. We need to choose what types of blocks to add in each stage. We increase the number of blocks with more parameters first. To compute how many blocks to add in a stage, we multiply the depth of the stage by depth coefficient and round the multiplication result. Width scaling is applied to all blocks equally. Finally, we consider latency when we scale.

\subsubsection{Add h-swish and SE}
In addition to compound scaling, we add two components in the post-processing step: \textit{h-swish} activation function and squeeze-and-excitation (SE) block. A recent study~\cite{park2020profit} reports that SE and the h-swish activation function are no hurdles for 8-bit quantization. They could quantize a network with SE and h-swish without noticeable accuracy loss. 

Extensive studies have been conducted to find a better activation function than ReLU, and the swish activation function~\cite{ramachandran2017searching} was found. Several neural networks~\cite{tan2019mixconv, mei2019atomnas, tan2019efficientnet} use swish activation function instead of ReLU to improve accuracy. Howard et al.~\cite{howard2019searching} proposed a quantization-friendly version of the swish activation function called h-swish that has a similar impact on accuracy. So, we replace ReLU with h-swish~\cite{howard2019searching} activation function. 

\begin{figure}[!ht]
  \centering
  \includegraphics[width=0.9\linewidth]{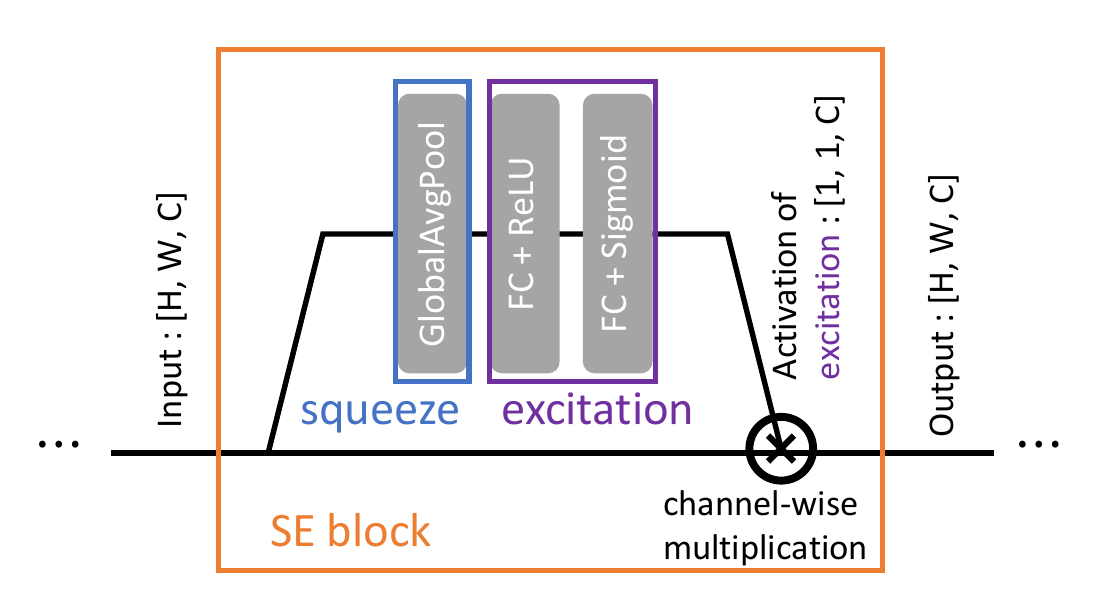}
  \caption{Structure of SE block}
  \label{fig:se}
\end{figure}

Squeeze-and-Excitation(SE) is a lightweight operation which is shown to be beneficial to accuracy~\cite{hu2018squeeze}. Figure~\ref{fig:se} depicts the structure of a SE block. For a given input feature map, it first computes the importance of the feature channels a representative value for global spatial information of each feature channel by global average pooling. After such \textit{squeeze} operation generates channel-wise statistics, \textit{excitation} operation captures channel-wise dependencies by two cascaded fully-connected layers to produce activation values, which represents the importance of each feature channel.
Finally, channel-wise multiplication is performed between the activation values induced by the excitation operation and the input feature map for each channel. SE block is used in many recent architectures~\cite{tan2019efficientnet, howard2019searching, radosavovic2020designing}. By adding SE blocks to the baseline network, we also observe the accuracy improvement.

\subsubsection{Selective removal of SE}\label{rmSE}

\begin{figure}[t]
  \centering
  \includegraphics[width=\linewidth]{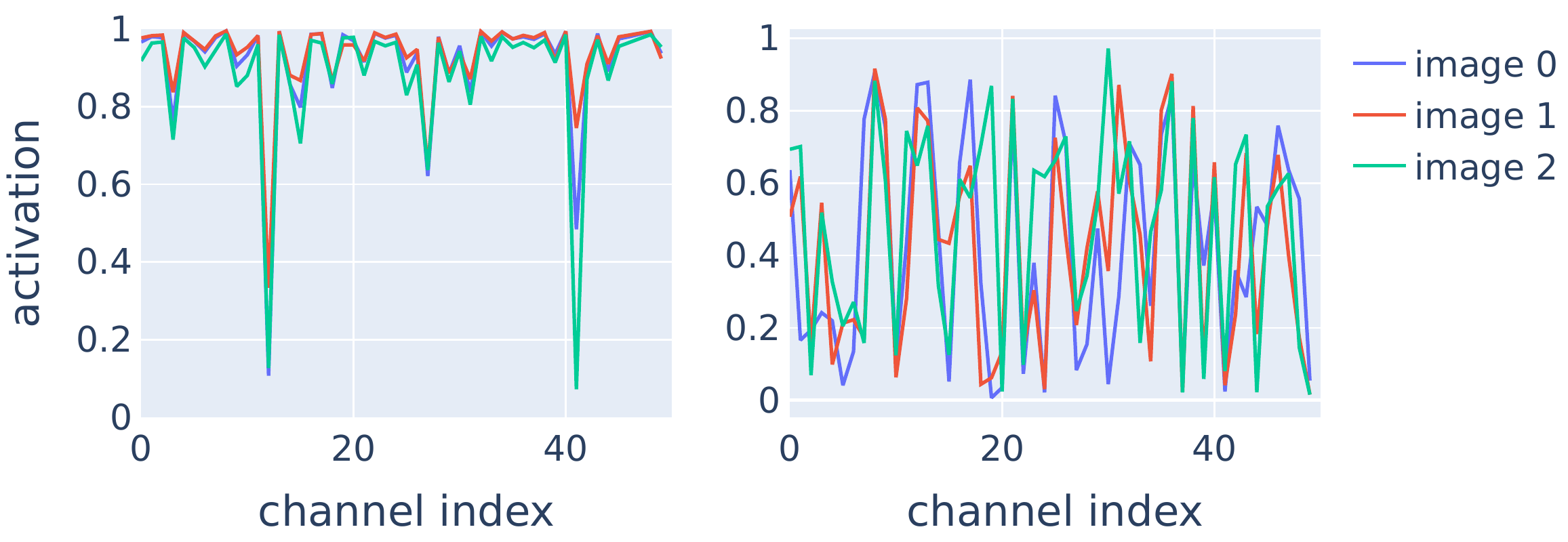}
  \caption{Distribution of activation values produced by two different SE blocks for three different images} %
  \label{fig:excitation}
\end{figure}

Figure~\ref{fig:excitation} depicts an example distribution of activation values produced by two different SE blocks for three different images. The authors of the original paper~\cite{hu2018squeeze} conjectured that if such distribution from a SE block does not differ widely between image classes, the SE block is not important. Thus, after training, they obtained averaged activation values of a SE block over multiple images in the same class.
They compared the distributions of the averaged values over different image classes. They observed that removing the SE blocks that have similar distributions over different image classes incurs only a marginal loss in accuracy.

Inspired by this observation, we propose to remove SE blocks selectively to minimize the additional computation cost caused by SE blocks.
We obtain activation values from a SE block for each input image and measure how the distribution of activation values varies over different input images. 
For each channel c, we calculate the standard deviation $\sigma_c$ of activation values over different images. If $\sigma_c$ is small in most channels, the activation values from the SE block does not differ much over images. Conceptually, it implies that the SE block does not help to discriminate further which channel is more influential. From the engineering perspective, it means that channel-wise multiplication of a SE block is similar to constant multiplication, which can be handled by the following convolutional layer. %

We define a metric as \textit{the average of standard deviation values $\sigma_c$ over all channels} that represent the diverseness of the activation distribution over different images. If the metric value is small, we remove the SE block. For example, in Figure~\ref{fig:excitation}, our metric of the SE block on the left side has a value of 0.021, while the right side has a value of 0.118, more than 5x larger than the left side; The left side is a better candidate for SE block removal. When we remove SE blocks according to this metric, the accuracy is found to be similar, while the latency got shorter (Table~\ref{tab:withexisting}).

\begin{figure*}[t]
  \centering
  \includegraphics[width=\linewidth]{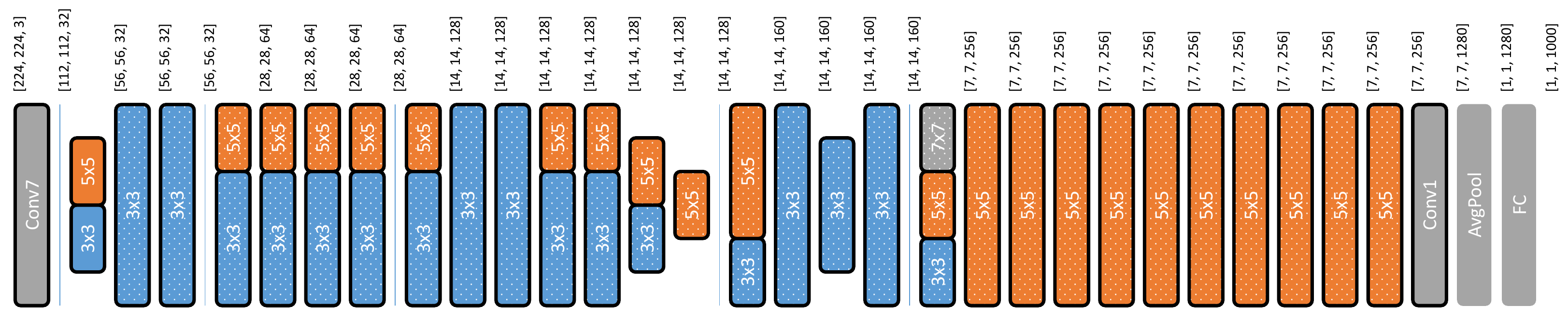}
  \caption{ours-M+ architecture. Height of each block in the picture is proportional to expansion ratio. SE-applied blocks are depicted as dotted blocks.}
  \label{fig:searchres}
\end{figure*}

\section{Experiments}

\subsection{Setup}

We evaluate the proposed NAS technique for image classification with the ImageNet dataset. The current implementation is made for MIDAP~\cite{kang2019novel} that can perform DWConv and SE operations efficiently so that MBConv is preferred to full 3-D convolution as the basic building block, as explained above.  %
Latencies on the target NPU are obtained with the cycle-accurate simulator\footnote{\url{https://github.com/cap-lab/MidapSim}}.

A superkernel has two parameters to search: expansion ratio and kernel size. To limit the search space, we choose the expansion ratio among 0, 2, 4, and 6, and the kernel size between 3 and 5 when MBConv or full convolution is used as the building block. In the case of the MixConv-based building block, we use $N$=3 superkenels whose expansion ratio is 0 or 2; The sum of the expansion ratio of three superkernels has the same range as the expansion ratio of a single MBConv block. To allow three superkernels to have different kernel sizes, we let one of three superkernels be able to have 7 as the kernel size. %

In the first phase of the neural architecture search, we train the supernet by randomly choosing one of the candidate subgraphs in each training step. We train the supernet for 8 epochs, with $\lambda_1=0$ in the loss function of Eq.~\ref{lossformula}, focusing only on the accuracy. We decrease the learning rate by 0.97 every 2.4 epochs, starting from 0.064. The other setting for network training is displayed in Table~\ref{tab:train_setting}. 
Gradient clipping with a value of 10 is used in this phase. In the second phase, we set $\lambda_1=15, \lambda_2=100$ to consider latency in the loss function, and optimize the weights and threshold values of supernet for 2 epochs. After this second phase finishes, the final architecture topology is decided.

Next, we train the final architecture again to determine the filter weights for 350 epochs with the ImageNet again, using the same setting described in Table~\ref{tab:train_setting}. Unlike the search phase, the learning rate is increased from 0 to 0.064 in the first 5 epochs, then decayed by 0.97 every 2.4 epochs.
Since we observed that the batch size is critical to accuracy when using the EfficientNet training code, we use a large batch size. %
Both network architecture search and final training are conducted on Google Cloud TPUs. %

\begin{table}[!ht]
\centering
\caption{Settings for network training, which is similar to~\cite{tan2019efficientnet} \protect\footnotemark.}
\label{tab:train_setting}
\resizebox{\linewidth}{!}{%
\begin{tabular}{|l|l|}
\hline
train batch size           & 1024                    \\ \hline
optimizer     & \begin{tabular}[c]{@{}l@{}}RMSProp with decay 0.9, \\ momentum 0.9, epsilon 0.001\end{tabular}                                   \\ \hline
image preprocessing        & Inception preprocessing \\ \hline
weight decay               & 1e-5                    \\ \hline
label smoothing            & 0.1                     \\ \hline
stochastic depth~\cite{huang2016deep}& 0.2           \\ \hline
Exponential Moving Average & 0.9999                  \\ \hline
\end{tabular}%
}
\end{table}
\footnotetext{The setting is similar to EfficientNet training code: \url{https://github.com/tensorflow/tpu/tree/master/models/official/efficientnet}}

\subsection{Supernet Design}\label{exp_proposed}

\begin{table}[!ht]
\centering
  \caption{The supernet architecture of the proposed NAS technique}
  \label{tab:supernet}
  \begin{tabular}{ccccc}
    \toprule
    input&block&width&depth&strides\\
    \midrule
    \begin{math}224^2 \texttimes 3\end{math}&7\texttimes7 conv&32&1&2\\
    \begin{math}112^2 \texttimes 3\end{math}2&TBD&32&\begin{math}d_1\end{math}&2\\
    \begin{math}56^2 \texttimes 32\end{math}&TBD&64&\begin{math}d_2\end{math}&2\\
    \begin{math}28^2 \texttimes 64\end{math}&TBD&128&\begin{math}d_3\end{math}&2\\
    \begin{math}14^2 \texttimes 128\end{math}&TBD&160&\begin{math}d_4\end{math}&1\\
    \begin{math}14^2 \texttimes 160\end{math}&TBD&256&\begin{math}d_5\end{math}&2\\
    \begin{math}7^2 \texttimes 256\end{math}&fc&1280&1&1\\
    \begin{math}7^2 \texttimes 1280\end{math}&avgpool&1280&1&1\\
    \begin{math}1^2 \texttimes 1280\end{math}&fc&1000&1&1\\
  \bottomrule
\end{tabular}
\end{table}

In the proposed NAS technique, two major extensions are made to the supernet, compared with the original SinglePath NAS technique. Table~\ref{tab:supernet} shows the proposed supernet architecture with configuration parameters, block types and depths.
It starts with a 7x7 convolution layer, followed by 5 stages that have a different number of blocks for feature extraction and 2 fully-connected networks for classification. %

The first extension is to allow stages to have a different number of blocks. To verify the goodness of this extension, we design two kinds of MBConv-based supernet with 20 blocks in total: a supernet with \textit{constant depth}(baseline), %
a supernet with \textit{linear depth} where the cumulative depth up to a specific stage is proportional to the width of the stage.

\begin{table}[!ht]
\centering
  \caption{Comparison of two MBConv-based supernets with different block assignment schemes to stages}
  \label{tab:depthwidth}
  \begin{tabular}{ccc}
    \toprule
    strategy&accuracy (\%)&latency (ms)\\
    \midrule
    constant (baseline) & 75.85 & 1.28\\
    \textbf{linear} & \textbf{77.09} & 1.24\\
  \bottomrule
\end{tabular}
\end{table}

As shown in Table~\ref{tab:depthwidth}, a supernet with linear depth outperforms a supernet with constant depth in terms of accuracy with similar latency. It confirms that this simple change of block assignment in supernet gives notable accuracy boost with the same latency constraint, without any additional optimization techniques. 

\begin{table}[!ht]
\centering
  \caption{Comparison of two supernets with different building blocks: MBConv and MixConv}
  \label{tab:MBvsMBConv}
  \begin{tabular}{ccc}
    \toprule
    building block&accuracy (\%)&latency (ms)\\
    \midrule
    DWConv (baseline) & 75.85 & 1.28 \\
    \textbf{MixConv} & \textbf{76.58} & 1.30 \\
  \bottomrule
\end{tabular}
\end{table}

The second extension is to use multiple parallel superkernels in an MBConv block. To verify the benefit of it, we compare two different supernets with the same number of blocks in each stage. The accuracy and latency performance of the baseline supernet is the same as the previous experimental result shown in Table~\ref{tab:depthwidth}. Table~\ref{tab:MBvsMBConv} shows that the extended supernet with MixConv-based building blocks gives a better accuracy-latency tradeoff.

\subsection{Comparison with existing models}

We apply the proposed NAS method with the supernet architecture described above. The depth of 5 stages is set to $3,4,7,4,11$, respectively. The latency constraint is set to 2.5 ms that corresponds to the latency of EfficientNet-B1 on our target NPU, MIDAP. Table~\ref{tab:withexisting} compares our search results with the state-of-the-art models: EdgeTPU~\cite{gupta2020accelerator}, EfficientNet~\cite{tan2019efficientnet}, Once-For-All~\cite{cai2019once}. The latency of the other models is obtained by running the network on the MIDAP cycle-accurate simulator. We compare the accuracy without quantization, assuming that quantization effects will be similar to all models. %

\begin{table*}[t]
  \caption{Performance comparison among network models with the ImageNet dataset. ($^{*}$ : Trained again with our training code. $^{**}$ : Trained again with official EfficientNet code using baseline preprocessing only. $^{\dagger\dagger}$ : We convert the wall clock time to \textit{GPU-hours}~\cite{stamoulis2020single} to compare with the other methods. + : using squeeze-excitation and the h-swish function. $^{\dagger}$ : When we compare the search cost, we compare the time needed to get one final architecture. $^{\ddagger}$: The search time contains training time.)}
  \label{tab:withexisting}
  \begin{tabular}{cccccc}
    \toprule
    Model &FP32 Top-1 acc(\%)&latency(ms)&\#Params&\#FLOPS&Search Cost(h)\\
    \midrule
    EdgeTPU-S~\cite{gupta2020accelerator} & 77.23 & 4.41& 5.4M & 2.35B & 40,000\\
    Inception V3~\cite{szegedy2016rethinking} & 78.8 & 6.75 & 23.8M & 5.71B & manual \\
    EfficientNet-B1~\cite{tan2019efficientnet} & 78.8 & 2.47 & 7.8M & 0.69B & 40,000\\
    EfficientNet-lite2~\cite{tan2019efficientnet} & 77.6 & 2.51 & 6.1M & 0.86B & 40,000\\
    random selection & 78.55$\sim$79.19 & 2.45$\sim$2.49 & 8.7M$\sim$11.0M & 0.97B$\sim$1.14B & 720(7200)$^{\ddagger}$\\
    random search & 78.93 & 2.49 & 10.4M & 1.06B & 3(30)$^{\dagger\dagger}$ \\
    ours-M & \textbf{79.35} & 2.47 & 12.8M & 1.29B & 3(30)$^{\dagger\dagger}$\\
    \midrule
    EdgeTPU-M~\cite{gupta2020accelerator} & 78.69 & 6.86 & 6.9M & 3.66B & 40,000\\
    EfficientNet-B2~\cite{tan2019efficientnet} & 79.8 & 3.31 & 9.1M & 0.99B & 40,000\\
    EfficientNet-lite3~\cite{tan2019efficientnet} & 79.8(79.15)$^{**}$ & 3.50 & 8.2M & 1.38B & 40,000\\
    Once-For-All~\cite{cai2019once} & 80.0(78.50)$^{*}$ & 2.18 & 9.1M & 0.60B &   1,200$^{\dagger}$\\
    ours-M+ & \textbf{80.28 }& 2.86& 15.4M & 1.29B & 3(30)$^{\dagger\dagger}$ \\
    \midrule
    EfficientNet-B3~\cite{tan2019efficientnet} & 81.0 & 5.22 & 12.2M & 1.83B & 40,000\\
    EfficientNet-lite4~\cite{tan2019efficientnet} & 81.5(80.38)$^{**}$ & 5.84 & 13.0M & 2.55B & 40,000\\
    ours-L+ & \textbf{81.49 }& 5.23& 20.7M & 2.47B & 3(30)$^{\dagger\dagger}$ \\
    \midrule
    EdgeTPU-L~\cite{gupta2020accelerator} & 80.62 & 15.55 & 10.6M & 9.66B & 40,000\\
    EfficientNet-B4~\cite{tan2019efficientnet} & 82.6 & 11.42 & 19.3M & 4.39B & 40,000\\
    ours-XL+ & \textbf{82.67 }& 11.87 & 27.9M & 5.95B & 3(30)$^{\dagger\dagger}$ \\
    ours-XL-rmSE+ & \textbf{82.72}& 11.66 & 26.9M & 5.95B & 3(30)$^{\dagger\dagger}$ \\
  \bottomrule
\end{tabular}
\end{table*}

As shown in Table~\ref{tab:withexisting}, the baseline model, \textit{ours-M}, found by the proposed NAS technique has higher accuracy than the other models on our target NPU; ours-M achieves more than 1.7\% higher top-1 accuracy than EfficientNet-lite2 with similar latency. Moreover, it is 0.5\% higher than EfficientNet-B1, even without using SE and h-swish activation function. Note that the number of parameters and the number of FLOPS in ours-M is larger than EfficientNet-B1. It implies that the complexity of the network is not a direct indicator of the end-to-end latency of the network. The end-to-end latency depends on the NPU architecture, and the proposed NAS technique could find a larger network with shorter latency by adding the latency factor to the loss function directly. The main benefit comes from different block assignment to stages.  

We improve the baseline network by adding the h-swish activation function and squeeze-and-excitation(SE) block to get the \textit{ours-M+} model. %
Figure~\ref{fig:searchres} shows the topology of ours-M+ architecture in which the height of each block is proportional to the expansion ratio of the block. Compared with the baseline network, ours-M, we achieve around 1\% accuracy boost with ours-M+, paying the cost of 16\% latency increase. %
This model outperforms the other models, 0.5\% higher accuracy and 14\% faster than EfficientNet-B2. Since EfficientNet-B2 is too large to run with the default configuration on MIDAP, we increase the memory size for filter weights. %

Next, we applied compound scaling~\cite{tan2019efficientnet} to ours-M+ to obtain \textit{ours-L+} and \textit{ours-XL+}. %
When we determine scaling coefficients, we keep the linear relationship 
between the cumulative depth and width of each stage, and scale the input image size more aggressively than~\cite{tan2019efficientnet}. %
We make the number of filters to be multiples of 16 to maximize the MAC unit utilization on MIDAP. When we train our scaled model, we set the dropout ratio to 0.4, similar to EfficientNet-B4 training.
The accuracy of ours-L+ is higher than EfficientNet-B3 and EfficientNet-lite4, while the accuracy of ours-XL+ is similar to EfficientNet-B4. 
Note that the difference between the searched network and the EfficientNet decreases as the network size increases.

Finally, we selectively removed SE blocks from ours-XL+, resulting in \textit{ours-XL-rmSE+}. We collected the activation values using randomly sampled 10K images from the training dataset and calculated the metric explained in Sec.~\ref{rmSE}.
After removing SE blocks from ours-XL+ based on the metric, only about 60\% of the blocks in the network have SE blocks. As a result, we could make the latency shorter, while the accuracy was slightly improved than ours-XL+. This model achieves 82.72\% top-1 accuracy with only 11.66ms latency.
It is much better than EfficientNet-EdgeTPU-L~\cite{gupta2020accelerator} that achieves 80.62\% FP32 top-1 accuracy with more than 20ms on EdgeTPU. Our architecture on MIDAP is about 2 times faster with 2.1\% higher accuracy.

Finally, we compare the search time. Since the TPU is faster than GPU, %
we report the wall clock time and the estimated GPU time (in parenthesis) that is 10 times longer than the wall clock time in the last column of Table~\ref{tab:withexisting} %
Our method takes \textbf{3 hours}, which is much faster than the other methods. Note that we compare the total time to get one architecture from scratch without trained weights. Once-For-All~\cite{cai2019once} would require only short fine-tuning time after a neural architecture is searched. %
In contrast, we need to train the network after a network architecture is found. It took 40 hours on TPUv3 to train ours-M+. %

\subsection{Ablation Studies}
\subsubsection{Comparison with Random Search}
While most NAS techniques are not compared with a random search method, the authors~\cite{li2019random} reported that a random search method is highly competitive. So we conducted an experiment to compare the proposed NAS technique with two random search methods, exploring the same search space defined by the supernet structure of ours-M. 
First, we designed a simple random search method that has the similar time complexity of the proposed technique. In this method, we randomly generate 15 models having a similar latency with ours-M, from the same search space. Then we train each of them for 1 epoch with cosine learning rate decay. After evaluating each of them, we choose the architecture with the topmost top-1 accuracy and fully train it. In the second method, called  random selection, we randomly generate 20 models having a similar latency with ours-M and train them fully and take the architecture with the highest top-1 accuracy. Since the random selection method performs search and training simultaneously, it is slower than the proposed technique by the number of randomly generated models.  

Comparison results are reported in Table~\ref{tab:withexisting}. It is confirmed that both random selection and random search are quite competitive, but noticeably inferior to ours-M in terms of accuracy.
In detail, the worst case of random selection showed 0.8\% lower accuracy than ours-M. The best performance obtained from 20 randomly generated models is 79.19\%, still lower than the accuracy of ours-M. 
Note that random search and random selection show similar performance that is no smaller than the other networks. It means that the search space defined by the supernet architecture has a more significant effect on the accuracy than the search method.

\subsubsection{Comparison between compound scaling and direct search}

\begin{table}[h]
\centering
  \caption{Comparison between compound scaling and direct search on large target latency.
  }
  \label{tab:CPScale}
  \begin{tabular}{ccc}
    \toprule
    & accuracy (\%) & latency (ms) \\
    \midrule
    Compound scale & 81.78 & 10.81 \\
    Direct search & 81.87 & 10.84 \\
  \bottomrule
\end{tabular}
\end{table}

There are two methods to find an architecture with a loose latency constraint. One is to use compound scaling that scales a small network with shorter latency, and the other is to search a network directly. To compare these two methods, 
we first scaled ours-M using the same scaling coefficients that we used to scale ours-M+ to ours-L+ and trained it. When conducting a direct search, we scaled the depth and width of the supernet and the input image size first and applied the proposed NAS technique for the scaled supernet. We used batch size 512 instead of 1024 during the architecture search due to the memory limitation of TPU. %
The comparison result is shown in Table~\ref{tab:CPScale} in terms of top-1 accuracy(\%) and the latency on the target NPU(ms). 
Two results were similar while direct search needed 10 hours on TPUv3; It means that compound scaling is an effective method to find a large network fast.

\subsubsection{Affect of SE and h-swish}

\begin{table}[h]
\centering
  \caption{Impact of SE and h-swish on accuracy.}
  \label{tab:SEHswish}
  \begin{tabular}{ccc}
    \toprule
    &ReLU&h-swish\\
    \midrule
    w/o SE & 79.35 & 79.36 \\
    w SE & 80.04 & 80.28 \\
  \bottomrule
\end{tabular}
\end{table}
To examine how SE and h-swish impact accuracy individually, we compare four combinations as displayed in Table~\ref{tab:SEHswish}. The baseline is ours-M that does not use SE and h-swish activation function. Replacing ReLU with h-swish gives a marginal improvement on accuracy while adding SE blocks improves the accuracy noticeably. %
Adding both SE and h-swish activation function improves the accuracy by around 1\%. %

\section{Conclusion}
In this work, we propose a fast NPU-aware NAS methodology extending the Single-Path NAS technique~\cite{stamoulis2019single}. We modify the supernet architecture by varying the number of blocks in stages and adding mixed depthwise convolution~\cite{tan2019mixconv} to the search space. By modifying the loss function to directly include the target latency estimated by a cycle-accurate simulator of the target NPU, we could find a better baseline architecture with a shorter latency than the latency constraint. Using a tight latency constraint, we can reduce the search space to find the baseline network fast. Afterward, we apply compound scaling to find a larger network than the baseline network, and add SE blocks and h-swish activation functions in the post-processing step. 
Through the proposed NAS methodology, we could obtain a network with 82.72\% accuracy with 11.66ms latency on our target NPU, without special data augmentation in training. It dominates the existing network models on the target NPU. It confirms the importance of supernet architecture design for a given NPU and effectiveness of the three-step approach in the proposed NAS methodology: supernet design, SinglePath NAS with a tighter latency constraint, and compound scaling and post-processing.

\begin{acks}
This work is supported by the National Research Foundation of Korea (NRF) grant funded by
the Korea government(MSIP) (NRF-2019R1A2B5B02069406). Also, we acknowledge support from the TensorFlow Research Cloud (TFRC) program.
\end{acks}

\bibliographystyle{ACM-Reference-Format}
\bibliography{ref}

\end{CJK}
\end{document}